\pdfoutput=1

\documentclass[11pt]{article}

\usepackage[]{ACL2023}

\usepackage{times}
\usepackage{latexsym}

\usepackage{array}
\newcolumntype{?}{!{\vrule width 1.5pt}}

\usepackage[T1]{fontenc}

\usepackage[utf8]{inputenc}

\usepackage{microtype}
\usepackage{amsmath}

\usepackage{svg}
\usepackage{csquotes, comment}
\usepackage{multirow}
\usepackage{xspace}
\usepackage{adjustbox}
\usepackage{booktabs}
\usepackage{tabularx,lipsum,environ,amssymb}    
\usepackage{bigdelim}
\usepackage{caption}
\usepackage{subcaption}
\usepackage{graphicx}
\usepackage{float}

\definecolor{mygreen}{rgb}{0, 0.8, 0}
\definecolor{myred}{rgb}{0.8, 0, 0}

\usepackage{inconsolata}
\usepackage{xcolor}

\newcommand{\mask}{\texttt{[MASK]}}
\newcommand{\x}{\texttt{[X]}}
\newcommand{\y}{\texttt{[Y]}}
\newcommand\subject{s\xspace}
\newcommand\relation{r\xspace}

\newcommand{\srp}{$\langle$\texttt{subject}, \texttt{relation}$\rangle$\xspace}
\newcommand{\srotupleexample}[3]{$\langle$\texttt{#1}, \texttt{#2}, \texttt{#3}$\rangle$\xspace}
\newcommand{\srpexample}[2]{$\langle$\texttt{#1}, \texttt{#2}$\rangle$\xspace}
\newcommand{\autoprompt}{A{\small UTO}P{\small ROMPT}}
\newcommand{\optiprompt}{O{\small PTI}P{\small ROMPT}}
\newcommand{\rel}[1]{\verb~#1~}

\title{Extracting Multi-valued Relations from Language Models}

\author{Sneha Singhania, Simon Razniewski, Gerhard Weikum \\
        Max Planck Institute for Informatics \\ 
        \texttt{\{ssinghan, srazniew, weikum\}@mpi-inf.mpg.de}}

\begin{document}
\maketitle
\begin{abstract}

The widespread usage of latent language representations via pre-trained language models (LMs) suggests that they are a promising source of structured knowledge. However, existing methods focus only on a single object per subject-relation pair, even though often multiple objects are correct. To overcome this limitation, we analyze these representations for their potential to yield materialized multi-object relational knowledge. We formulate the problem as a \emph{rank-then-select} task. For \emph{ranking} candidate objects, we evaluate existing prompting techniques and propose new ones incorporating domain knowledge.
Among the \emph{selection} methods, we find that choosing objects with a likelihood above a learned relation-specific threshold gives a 49.5\% F1 score.
Our results highlight the difficulty of employing LMs for the multi-valued slot-filling task, and pave the way for further research on extracting relational knowledge from latent language representations.

\end{abstract}

\section{Introduction}
\citet{petroni-etal-2019-language} showcased the potential of relation-specific probes for extracting implicit knowledge from latent language representations. But the viability of materializing factual knowledge directly from LMs
remain open problems \cite{Razniewski2021LMsKBs, AlKhamissi2021LMsKBsSurvery}.

\begin{figure}[t]
\centering
\includegraphics[width=\linewidth, 
trim={0cm 7.1cm 14cm 0cm},
clip=true]{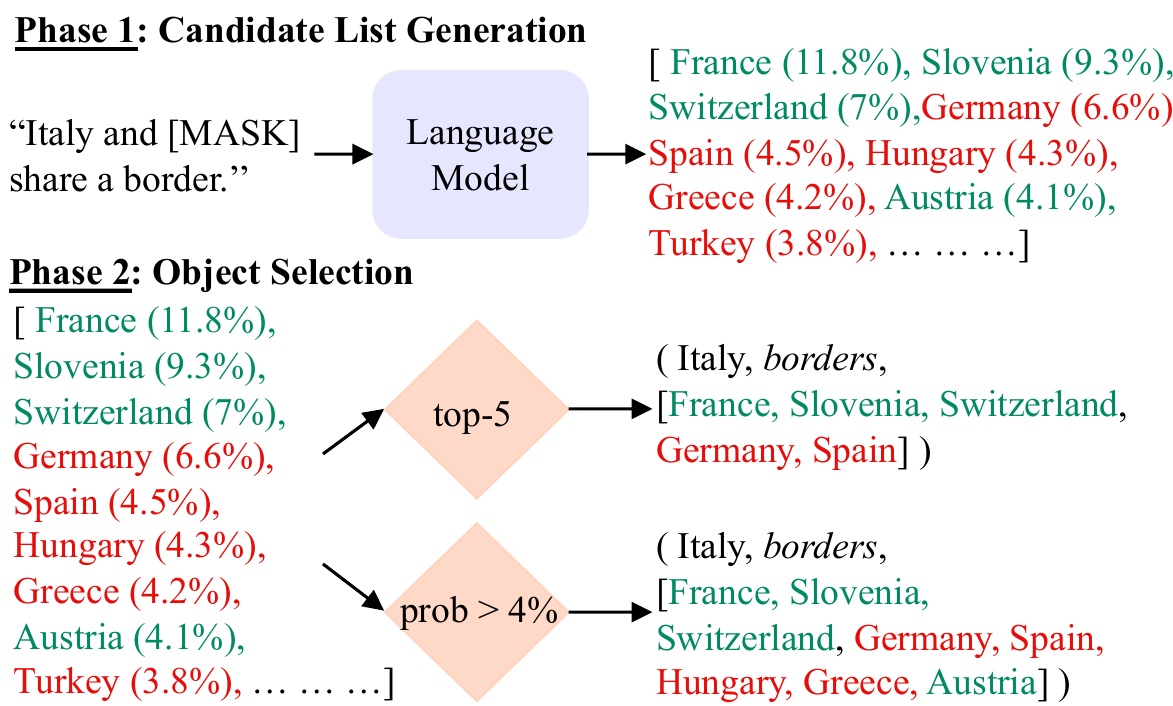}
\caption{Probing LMs to extract objects for multi-valued relations.
}
\label{fig:problemstatement}
\end{figure}

Building upon the LAMA framework, where an LM 
predicts an object in the slot for
given a cloze-style prompt such as ``Dante was born in \mask'', several methods \cite{jiang-etal-2020-know, shin-etal-2020-autoprompt, zhong-etal-2021-factual, qin-eisner-2021-learning} design effective prompts for factual information extraction. Importantly, however, all these methods implicitly assume the existence of a single correct object per (subject, relation)-pair and evaluate on precision at rank 1. In reality, many relations have multiple correct values
as illustrated in Figure~\ref{fig:problemstatement}.

In this paper, we focus on probing LMs directly for multi-valued slot-filling task.
In a zero-shot setting, an LM is probed using prompts that include \textit{a subject and a multi-valued relation} to generate a list of candidate objects. However, prior knowledge of the no. of correct objects is generally unknown. Since the LM's probabilities alone do not provide a clear indication of the objects' factual accuracy~\cite{jiang-etal-2021-know, holtzman-etal-2021-surface}, we apply various selection mechanisms on the generated list to choose the correct objects.

We probe LMs using existing prompting techniques and introduce new relation-specific prompts. The generated object lists are evaluated by their order, and our manual prompts result in higher-quality lists than the state-of-the-art automated methods. Our prompts outperform the best baseline, SoftPrompts, by ca. 5\% points on the three most challenging relations (parts of chemical compounds, official languages of countries, instruments of musicians), while being competitive on the other four relations. While evaluating the output of selection mechanisms, our best approach achieves 54.1\% precision, 50.8\% recall, and 49.5\% F1 score.
These trends in scores demonstrate the difficulty of extracting complete knowledge from internal LMs representations.

\section{Background}

The idea of knowledge extraction using LMs representations was put forward by \citet{Radford2019LanguageMA, petroni-etal-2019-language}, and has since received much attention. Several approaches~\cite{Bouraoui2019InducingRK, goswami-etal-2020-unsupervised-relation, KnowPrompt} use relational metadata and textual corpora to tune the LM using knowledge-enriched representations and efficiently extract knowledge. \citet{jiang-etal-2020-x, kassner-etal-2021-multilingual} focus on multilingual knowledge extraction, while \citet{dhingra-etal-2022-time} looks at extracting temporal knowledge from LMs.
Current methods usually sample from correct objects for a given subject-relation pair by employing the hits@k metric and do not enforce deliberate accept/reject decisions on the outputs.

\section{Methodology}
\paragraph{Candidate List Generation}
When probing an LM using a cloze-style prompt, the LM generates a probability distribution over the vocabulary tokens to fill in the masked position. For our task, we use prompts mentioning a subject-relation \srpexample{\subject}{\relation} pair and consider the resulting ranked list of tokens (w/ their corresponding probability scores) as candidates. In a zero-shot setting, the LM is probed using two types of prompts: (i) \textit{discrete prompts} such as LPAQA~\cite{jiang-etal-2020-know} and \autoprompt~\cite{shin-etal-2020-autoprompt}, and (ii) \textit{continuous prompts} such as \optiprompt~\cite{zhong-etal-2021-factual} and SoftPrompts~\cite{qin-eisner-2021-learning}. 

In addition, we propose a collection of carefully designed manual prompts to evaluate and compare the generated objects for multi-valued relations. We create  50 diverse relation-specific prompts by incorporating domain knowledge, relation type, and variations in sentence structure and grammar. Our prompts differ in verb form, tense, placement of the masked token (whether it is predicted in a prefix, suffix, or cloze-style), and whether or not there is a period and object type in the context.

\paragraph{Selection Mechanisms}
We experiment with the following parameterized mechanisms on the generated object lists to get a valid subset of triples.\\
\textbf{Top-$k$}: The most probable $k$ objects are selected.\\
\textbf{Prob-$x$}: Objects with a probability greater than or equal to $x$ are chosen.\\
\textbf{Cumul-$x$}: Retain all objects, in order of probability, whose summed probability is no larger than $x$. In difference with Prob-$x$, it would enable retaining candidates of similarly moderate probability. \\
\textbf{Count Probe}:
We probe the LM again to get an object count prediction for a \srpexample{\subject}{\relation}. A count probe could be as follows, \textit{``Italy borders a total of \mask{} countries''}. From the list of generated tokens, the highest-ranked integer type token (either in alphabetical or numerical form) is used to subset the original object list.\\
\textbf{Verification Probe}: We probe the same LM again on each candidate object to factually verify the generated subject-relation-object \srotupleexample{s}{r}{o} triple. A verification probe could be as follows, \textit{``Italy and France share a border? Answer: \mask''}. We compare the relative probabilities of the \texttt{\_yes} versus \texttt{\_no} tokens in the masked position, using $(p_{\texttt{\_yes}}-p_{ \texttt{\_no}} > \alpha)$, with $\alpha$ is a hyper-parameter, to determine the correctness of the original candidate object, France.  All the original candidate objects satisfying the comparison condition are selected.

\section{Experiment}
\paragraph{Dataset} The seven diverse multi-valued relations appearing in LAMA benchmark~\cite{petroni-etal-2019-language} are chosen. For each relation, approx. $200$ subjects and a complete list of objects from the popular Wikidata KB~\cite{Vrandei2014WikidataAF} are sampled. The subjects were picked based on popularity, measured using Wikidata ID and count of Twitter followers for person-type subjects\footnote{\href{https://github.com/snehasinghania/multi_valued_slot_filling}{https://github.com/snehasinghania/multi\_valued\_slot\_filling}}.

\paragraph{Evaluation} In the \textit{ranking} phase, the quality of a candidate list is assessed using the maximally possible F1 score, defined as the highest possible F1 score achieved by applying the top-$k$ selection mechanism with the optimal $k$ value. The optimal $k$ is found by iterating over all possible choices of $k$ and calculating the respective F1 score on the subset of candidate objects and ground-truth objects. In the \textit{selection} phase, the output triples obtained after applying a selection mechanism are evaluated using precision, recall, and F1 score.

\setlength{\tabcolsep}{1.5pt}
\begin{table}[t]
\resizebox{\columnwidth}{!}{ 
\centering
\begin{tabular}{lcccccc}
\toprule
\srp & Ours & Mined & Para & A{\small UTO} & O{\small PTI} & Soft\\
\toprule
compound, \textit{has-parts} &  \textbf{78.5} & 	38.1 & 29.5 & 51.4 & 68.1 & 71.6 \\ 
country, \textit{borders} & 72.8 & 	66.4 & 	64.9 & 71.6 & 	73.2 & \textbf{75.6} \\ 
country, \textit{official-lang} & \textbf{83.6} & 81.9 & 75.2 & 71.8 & 75.9 & 79.9 \\ 
person, \textit{instrument} & 62.5 & 	\textbf{63.4} & 61.3 & 	61.7 & 52.7 & 57.0 \\ 
person, \textit{speaks-lang} & \textbf{72.8} & 	69.1 & 41.1 & 52.8 & 71.5 & 69.0 \\ 
person, \textit{occupation}  & 33.2 & 40.2 & \textbf{44.9} & 37.5 & 36.6 & 36.9 \\ 
state, \textit{borders} & 25.7 & 23.8 & 24.3 & 24.4 & 25.9 & \textbf{25.9} \\ 

\midrule
Overall (avg.) & \textbf{61.3} & 54.7 & 	48.7 & 	53.0 &  57.7 & 59.4 \\
\bottomrule
\end{tabular}}
\caption{Comparing generated object lists by probing BERT-large using macro-averaged max-F1(\%) scores. Mined is LPAQA's mining-based prompts and Para is LPAQA's paraphrasing-based prompts. A{\small UTO} uses the open-sourced \autoprompt{} templates. O{\small PTI} and Soft are \optiprompt{} and SoftPrompts obtained by training the LM using author-released data and code. 
}
\label{tab:clg}
\end{table}

\paragraph{Setup} We reuse the best prompts reported by each prompting baseline, which are tuned on much larger data.
We probe BERT~\cite{devlin-etal-2019-bert} on each \srpexample{s}{r} and generate the $500$ most probable candidate objects. The generated list is post-processed to remove stopwords and other type-irrelevant objects depending on the relation type, only to retain sensible candidate objects. Our dataset is split into train, dev, and test, with 100/$\sim$50/50 subjects per relation, for tuning and estimating parameters in the selection mechanisms.

\section{Results}
\paragraph{Candidate List Generation}
We compare the candidate list generated by each prompting method in Table~\ref{tab:clg}. Our prompts generate the best object lists in terms of macro-averaged max-F1 score. Unlike our prompts, discrete prompts have a lower performance, while continuous counterparts have a similar performance. Surprisingly, \optiprompt{} obtained by initializing its continuous vectors using our prompts has a lower score.

\begin{table*}[htbp]
\resizebox{\textwidth}{!}{ 
\centering
\begin{tabular}{llclc}
\toprule
\srp & Our Prompts with best precision@1 & hits@1 & Our Prompts with best max-F1 & max-F1 \\ %
\midrule
compound, \textit{has-parts} & \x{} contains \mask{} atom & 78.50 & \x{} has \mask, which is an atom. & 78.52 \\ %
country, \textit{borders} & \x{} and \mask{} share a border& 	84.86 & \x{} and \mask{} share a border.& 72.82 \\ %
country, \textit{official-lang} & People of \x{} mostly speak in \mask. & 	93.37 & 	\mask{} is the main language of \x. & 	83.57 \\ %
person, \textit{instrument} & Musician \x{} plays \mask. & 	67.50 & 	\x{} plays \mask{}, which is an instrument & 62.45 \\ %
person, \textit{speaks-lang} & In which language can \x{} talk? Answer: \mask. & 	92.50 & 	\x{} speaks in \mask.& 72.78 \\ %
person, \textit{occupation} & \x{} is a well-known \mask. & 	59.00 & 	\x{} is a well-known \mask{} & 	33.21 \\ %
state, \textit{borders} &\mask, which is a \y, borders \x. & 	37.50 & 	\x{} and \mask{} share a border & 	25.71 \\ %
\bottomrule
\end{tabular}}
\caption{Our best prompts among the $50$ relation-specific prompts on precision@1 (\%) and max-F1 (\%). The \y{} slot takes the object-type information, e.g., in $\langle$state, \textit{borders}$\rangle$, \y{} could be ``state'', ``governate'', ``prefecture'', etc.}
\label{tab:metric-comparison}
\end{table*}

To validate the effectiveness of our prompts and inspect if optimizing prompts on precision@1 is sufficient for extraction on multi-valued relations, we compare the best prompts in terms of precision@1 and max-F1 score. In Table~\ref{tab:metric-comparison}, we see that prompts performing well on precision@1 are not necessarily the best on max-F1. Overall, prompts suitable for multi-valued extraction are more often of the prefix type, while prompts in the single-object case show more variance.
In Appendix, Table~\ref{tab:list-with-selection} shows examples of generated lists, and Tables~\ref{tab:chemicalcompound}-\ref{tab:state_borders} lists all the prompt templates. The large gap between precision@1 and max-F1 indicates the difficulty in designing task-specific prompts and raises the need for developing more robust prompts.

\setlength{\tabcolsep}{4pt}
\begin{table*}[t]
\resizebox{\textwidth}{!}{ 
\centering
\begin{tabular}{l|ccc|ccc|ccc|ccc|ccc|c}
\toprule
\multirow{2}{*}{\srp} & \multicolumn{3}{c|}{top-$k$} & \multicolumn{3}{c|}{prob-$x$}  & \multicolumn{3}{c|}{cumul-$x$}  & \multicolumn{3}{c|}{count-probe}  & \multicolumn{3}{c|}{verify-probe} & \multirow{2}{*}{max-F1} \\
\cline{2-16}
& p & r & f1 & p & r & f1 & p & r & f1 & p & r & f1 & p & r & f1 & \\ 
\hline
compound, \textit{has-parts} & 62.5 & 78.1 & 68.0 
                             & 60.3 & 76.4 & 65.4 
                             & 37.8 & 74.7& 37.8  
                             & 54.2 & 78.7 & 61.8 
                             & 16.1 & 70.7 & 22.9
                             & 78.5\\

country, \textit{borders} &  64.0 & 55.4 & 54.2
                                & 63.4 & 	58.7 & 	55.4
                                & 56.0 & 	58.4 & 	46.5
                                & 21.9 & 	71.7 & 	30.5
                                & 1.9 & 	68.2 & 	3.6 
                                & 72.8\\

country, \textit{official-lang}     & 96.0 & 74.1 & 80.1
                                    & 94.0 & 75.8 & 80.5
                                    & 52.8 & 71.3 & 43.2
                                    & 28.9 & 82.3 & 40.5
                                    & 27.0 & 32.4 & 4.9
                                    & 83.6\\ 

person, \textit{instrument} & 46.0 & 42.3 & 38.8
                                  & 51.7  & 	40.8 & 	39.1
                                  & 51.8 & 	41.4 & 	33.8
                                  & 18.2  & 	60.9 & 	25.5
                                  & 7.1 & 	24.4 & 	4.8
                                  &	62.5\\     

person, \textit{speaks-lang} & 52.5 & 59.8 & 55.1
                                 & 69.6 & 56.7 & 60.0
                                 & 56.2 & 57.3 & 46.8
                                 & 37.4 & 69.0 & 47.3
                                 & 3.5 & 53.0 & 6.4
                                 & 72.8 \\  

person, \textit{occupation} & 33.3 & 23.5 & 27.3
                            & 3.2 & 	85.1 & 	6.1
                            & 30.1 & 	36.2 & 	18.6
                            & 23.1 & 	30.0 & 	25.9
                            & 5.5 & 	41.7 & 	9.0
                            & 33.2\\

state, \textit{borders} &  24.4 & 22.6 & 22.9
                        & 63.1 & 	21.1 & 	24.9
                        & 	21.9 & 	18.3 & 	13.7
                        & 	10.0 & 	24.3 & 	13.9
                        & 	2.4 & 	26.2 & 	4.3 
                        & 25.7\\ 
\midrule

Overall (averaged) & 54.1 & 50.8 & \textbf{49.5} 
                   & \textbf{57.9}	& 59.2 & 47.4
                   & 	43.8 & 51.1 & 34.4 
                   & 	27.7 & \textbf{59.5} & 35.1
                   & 	9.1 & 45.3 & 8.0 
                   & 61.3 \\                    
\bottomrule

\end{tabular}
}
\caption{Results on comparing triples using precision, recall, and F1 score when probing BERT-large and applying a selection mechanism. The bold-faced numbers are the highest achieved precision, recall, and F1 scores.}
\label{tab:zero-shot}
\end{table*}

\paragraph{Object Selection} 
The candidate objects retained after applying a selection mechanism are compared against the ground-truth objects, and the results are shown in Table~\ref{tab:zero-shot}. The top-$k$ method achieves the best overall F1 score, which is the macro-average of individual \srpexample{s}{r} tuple-specific F1 scores. The individual F1 scores and max-F1 (upper bound) have a large gap since the probabilities of predicted tokens are not calibrated enough to match the actual factuality of the \srotupleexample{s}{r}{o} triple. Table~\ref{tab:cutoffs} in Appendix gives the prompt templates and learned parameters of each selection mechanism.

\section{Discussion}\label{sec:discussion}
Although BERT was probed for $500$ objects when generating object lists, only $119.7$ objects were retained after post-processing. Objects with invalid types occur due to the zero-shot setting. Also, other eminent masked LMs, including BERT-base, RoBERTa-base, and RoBERTa-large, achieve 60.61\%, 54.82\%, and 58.90\% max-F1 scores.

The max-F1 scores in Table~\ref{tab:clg} \&~\ref{tab:metric-comparison} are far from 100\%, i.e., LMs do not generate candidate lists that correctly rank all true objects above the false ones. In particular, max-F1 will not reach 100\% when correct objects are ranked too low or absent. We found that 26.90\% of valid objects in a candidate list were ranked below the optimal threshold and 27.75\% of valid objects were not generated at all.

In Table~\ref{tab:zero-shot}, the top-$k$ and prob-$x$ achieve balanced precision and recall scores. The count-probe achieves a high recall since almost always a count greater than 10 is predicted, and in our dataset, the average count of ground-truth objects across all \srpexample{s}{r} is in [1,10] range. 
In the verification probe, the parameter $\alpha$ is near zero for most relations, and the probability of \texttt{\_yes} is greater than \texttt{\_no}, leading to a selection of all the candidate objects. Although \citet{schick-schutze-2021-exploiting} and others show the effect of verbalizing labels to \texttt{\_yes} and \texttt{\_no} tokens in the few-shot setting on classification and inference tasks, optimally using them for factual knowledge extraction remains an open challenge.

\paragraph{Effect of Prompt Template}
In Table~\ref{tab:zero-shot}, each mechanism is evaluated on the candidate list generated using prompt templates shown in max-F1 column in Table~\ref{tab:metric-comparison}. However, by choosing a different set of prompts, a higher overall F1 score of 51.3\% with a lower 59.9\% max-F1 can be achieved by using the prob-$x$ method. Table~\ref{tab:zero-shot-f1score} in Appendix gives more details. 
This change in F1 scores shows the hardness of designing robust prompts.

\paragraph{Effect of Relation Type}
The candidate lists generated for popular subjects tend to have higher precision and recall. For instance, F1 score for (state, \textit{borders}) with top-$k$ is the lowest due to the presence of long-tail subjects. Also for a large possible set of valid objects, e.g., in \textit{occupation}, LM only generates common professions with a high probability and negatively impacts the F1 scores. This behavior, however, helps in \textit{language} type relations. A similar difference in performance can be observed in \citet{shin-etal-2020-autoprompt, zhong-etal-2021-factual}.  

\paragraph{Calibrate using Web Signals}
We used the query hit rate from Bing to calibrate and select objects from the candidate list. Bing receives each \srotupleexample{s}{r}{o} triple in the form of a natural language query.
The hit rate is used in two ways: (i) subset, objects with non-zero hit rate, (ii) rerank, objects with non-zero hit rate are calibrated to the highest probability. In contrast to prob-$x$ F1 scores, we observe a high increase in precision and decrease in recall applying (I) with a lower overall F1 score of 34.7\%, while the (II) method achieves higher recall and lower precision with a similar overall F1 score of 45.7\%. Table~\ref{tab:bing-based-calibration} in Appendix shows all the scoresx.

\paragraph{Effect of LM Size} We probed larger LMs like T5~\cite{t5model} and BART~\cite{lewis-etal-2020-bart}, which can generate a list of tokens with likelihoods using Beam search decoding algorithm, similar to masked LMs. With top-$k$ method, T5-large achieves 43.6\% precision, 41.7\% recall, and 40.3\% F1 score. BART-large achieves an even lower 32.0\% precision, 34.3\% recall, and 30.8\% F1 score. Table~\ref{tab:llm-results} in Appendix gives all the scores. 
These models tend to generate common objects and exhibit repetitive behavior. Also, the current trend is towards using autoregressive models like \mbox{(chat-)}GPT, like \citet{kalo,geva} used to extract multi-valued relations. However, unlike in our method, with no control over the selection mechanism, the LM directly outputs one final list in both works. While internally autoregressive models also use token probabilities that could be used for our approach, once one generates a full list, previously generated list items conflate the probabilities of items.

\section{Conclusion}
In this work, we evaluate using LM's internal representation for materializing factual knowledge on multi-valued relations. We utilize existing prompt engineering techniques and propose new prompts tailored for multi-valued relations.
The suggested selection mechanism approaches help to filter out valid triples. Our detailed analysis of the model's performance highlights the limitations of using zero-shot probing for the multi-valued slot-filling task. Future work could aim to improve overall precision and recall and measure the impact of using LMs to fill gaps in KBs.

\bibliography{anthology, ref}
\bibliographystyle{acl_natbib}

\appendix
\section{Appendix}\label{sec:appendix}
\begin{table*}[htbp]
\centering
\resizebox{\textwidth}{!}{ 
\begin{tabular}{l|l|r}
\toprule
\srp & Generated Object List & |GT|\\
\hline
(compound, \textit{has-parts}) & & \\
Calcium Carbonate & \textcolor{mygreen}{carbon    (0.47)}, \textcolor{myred}{hydrogen    (0.03)}, \textcolor{mygreen}{oxygen    (0.02)}, \textcolor{mygreen}{calcium    (0.01)}, \textcolor{myred}{silicon    (0.01)}, \textcolor{myred}{nitrogen    (0.002)}, \textcolor{myred}{sulfur    (0.002)} & \multicolumn{1}{c}{3}\\

Dopamine & \textcolor{mygreen}{hydrogen    (0.09)}, \textcolor{mygreen}{nitrogen    (0.05)}, \textcolor{mygreen}{carbon    (0.05)}, \textcolor{mygreen}{oxygen    (0.05)}, \textcolor{myred}{calcium    (0.02)}, \textcolor{myred}{sodium    (0.011)}, \textcolor{myred}{sulfur    (0.009)} & \multicolumn{1}{c}{4}\\

Sodium Chloride & \textcolor{myred}{carbon    (0.15)}, \textcolor{myred}{hydrogen    (0.09)}, \textcolor{myred}{oxygen    (0.03)}, \textcolor{myred}{silicon    (0.01)}, \textcolor{myred}{nitrogen    (0.01)}, \textcolor{myred}{sulfur    (0.006)}, \textcolor{mygreen}{sodium    (0.006)}& \multicolumn{1}{c}{2}\\

Thiocyanic Acid & \textcolor{mygreen}{hydrogen    (0.1)}, \textcolor{mygreen}{carbon    (0.1)}, \textcolor{myred}{oxygen    (0.03)}, \textcolor{mygreen}{nitrogen    (0.02)}, \textcolor{mygreen}{sulfur    (0.01)}, \textcolor{myred}{silicon    (0.003)}, \textcolor{myred}{sodium    (0.002)}& \multicolumn{1}{c}{4}\\

Water & \textcolor{mygreen}{oxygen    (0.17)}, \textcolor{mygreen}{hydrogen    (0.17)}, \textcolor{myred}{carbon    (0.05)}, \textcolor{myred}{nitrogen    (0.02)}, \textcolor{myred}{sodium    (0.01)}, \textcolor{myred}{mercury    (0.004)}, \textcolor{myred}{sulfur    (0.003)}& \multicolumn{1}{c}{2}\\

\hline
(country, \textit{borders}) & & \\
\bottomrule
\end{tabular}
}
\caption{Samples of generated object list for five unique subjects on multi-valued relations. The green highlighted valid objects, while the red ones are wrong. The |GT| column gives the total no. of ground-truth objects for the corresponding subject.}
\label{tab:list-with-selection}
\end{table*}

\setlength{\tabcolsep}{3pt}
\begin{table*}[htbp]
\centering
\resizebox{\textwidth}{!}{ 
\begin{tabular}{llcl}
\toprule
\srp & Metric & avg-cutoff & \multicolumn{1}{c}{Our Prompts} \\
\midrule
\multirow{5}{*}{compound, \textit{has-parts}} & 
top-$k$ & 4 & \hspace{-0.4em}\rdelim\}{3}{*}[\x{} has \mask, which is an atom] \\
& prob-$x$ & 0.02 & \\
& cumul-$x$ & 0.53 & \\
& count-alpha & 2.26 & \x{} consists of \mask{} elements.\\
& count-num & 4.64 & \x{} consists of \mask{} elements\\
& verify-probe & $\alpha=0.06$ & \x{} consists of \y{} atom. Is this correct? Answer: \mask.\\
\midrule

\multirow{5}{*}{country, \textit{borders}} & 
top-$k$ & 3 & \hspace{-0.4em}\rdelim\}{3}{*}[\x{} and \mask{} share a border.] \\
\bottomrule
\end{tabular}}
\caption{The prompt templates used for generating the object list. The avg-cutoff shows the learned parameters of each selection mechanism averaged across all subjects.}
\label{tab:cutoffs}
\end{table*}


\setlength{\tabcolsep}{2.5pt}
\begin{table*}[htbp]
\resizebox{\textwidth}{!}{ 
\centering
\begin{tabular}{llcccc|c}
\toprule
\srp & Our Prompts & avg-cutoff & Precision & Recall & F1 score & Max-F$_1$ \\
\hline
compound, \textit{has-parts} & \x{} has \mask, which is an atom. & 0.02	& 60.33 & 76.40	& 65.41 & 78.52 \\
country, \textit{borders} & \x{} has borders with \mask. & 0.07 & 74.73 & 55.41 & 58.45 & 71.41\\
country, \textit{official-lang} & The official language of \x{} is \mask. & 0.15 & 94.33 & 79.97 & 83.38 & 83.54 \\
person, \textit{instrument} & \x{} likes to play the \mask. & 0.12 & 51.33 & 46.52 & 41.48 & 58.19 \\
person, \textit{speaks-lang} & \x{} speaks in \mask.	& 0.05 & 69.60 & 56.72 & 59.99 & 72.78 \\
person, \textit{occupation} & \x{} is a \mask. & 0.01 & 22.79 & 31.26 & 25.42 & 29.30 \\
state, \textit{borders} & \x{} and \mask{} share a border & 0.04 & 63.10 & 21.12 & 24.91 & 25.71\\
\hline
Overall &  & & 62.32 & 52.49 & 51.29 & 59.9\\
\bottomrule
\end{tabular}}
\caption{Higher F1 score achieved by using a different set of our proposed prompts and prob-$x$ mechanism.}
\label{tab:zero-shot-f1score}
\end{table*}

\begin{table*}[t]
\resizebox{\textwidth}{!}{ 
\centering
\begin{tabular}{llccc|ccc|ccc}
\toprule
\multirow{2}{*}{\srp} & \multirow{2}{*}{Our Prompts}  & \multicolumn{3}{c}{Prob-$x$} & \multicolumn{3}{c}{Nonzero Hit-rate} & \multicolumn{3}{c}{ReRank}\\
\cmidrule(lr){3-5} \cmidrule(lr){6-8} \cmidrule(lr){9-11} 
& & p & r & f1  & p & r & f1   & p & r & f1 \\
\hline
compound, \textit{has-parts} & \x{} has \mask, which is an atom. 
                                                     & 60.3 & 76.4	& \textbf{65.4} 
                                                     & 84.2 & 17.7 & 17.0
                                                     & 57.0 & 80.9 & 64.1\\
country, \textit{borders} & \x{} has borders with \mask. 
                                               & 74.7 & 55.4 & 58.5 
                                               & 82.3 & 53.5 & 56.0
                                               & 58.9 & 69.4 & \textbf{58.8} \\
country, \textit{official-lang} & The official language of \x{} is \mask. 
                                                   & 94.3 & 80.0 & \textbf{83.4} 
                                                   & 91.7 & 68.3 & 71.1
                                                   & 69.3 & 81.6 & 69.3 \\
person, \textit{instrument} & \x{} likes to play the \mask. 
                                                 & 51.3 & 46.5 & \textbf{41.5}
                                                 & 83.0 & 30.1 & 33.8
                                                 & 27.7 & 53.2 & 32.4\\
person, \textit{speaks-lang} & \x{} speaks in \mask. 
                                                & 69.6 & 56.7 & \textbf{60.0} 
                                                & 83.0 & 19.7 & 22.1
                                                & 48.5 & 66.3 & 51.2\\
person, \textit{occupation} & \x{} is a \mask.
                                           & 23.0 & 31.3 & \textbf{25.4}
                                           & 19.7 & 23.7 & 20.3
                                           & 19.8 & 24.1 & 20.5\\
\bottomrule
\end{tabular}
}
\caption{Results on calibrating object probabilities with Bing hit rates}
\label{tab:bing-based-calibration}
\end{table*}

\begin{table*}[t]
\resizebox{\textwidth}{!}{ 
\centering
\begin{tabular}{llccc|ccc}
\toprule
\multirow{2}{*}{\srp} & \multirow{2}{*}{Our Prompts}  & \multicolumn{3}{c}{T5-large} & \multicolumn{3}{c}{BART-large} \\
\cmidrule(lr){3-5} \cmidrule(lr){6-8} 
& & p & r & f1   & p & r & f1 \\
\hline
compound, \textit{has-parts} & \x{} has \mask, which is an atom. & 67.3 & 64.9 & 61.5 & 58.7 & 56.2 & 56.2\\
country, \textit{borders} & \x{} and \mask{} share a border. & 43.8 & 53.1 & 45.0 & 44.8 & 60.4 & 47.3\\
country, \textit{official-lang} & \mask{} is the main language of \x. & 82.0 &	61.1 & 66.8 & 0 & 0 & 0 \\
person, \textit{instrument} & \x{} plays \mask, which is an instrument & 13.3 & 14.8 & 12.6 & 40.0 & 36.1 & 33.5\\
person, \textit{speaks-lang} & \x{} speaks in \mask. & 47.0 & 40.4 & 43.0 & 22.1 & 28.7 & 24.5\\
person, \textit{occupation} & \x{} is a well-known \mask & 27.2 & 36.4 & 30.8 & 36.5 & 37.1 & 34.4\\
state, \textit{borders} & \x{} and \mask{} share a border & 24.6 & 21.5 & 22.3 & 21.5 & 21.4 & 19.8\\
\midrule
Overall & & 43.6 &	41.7 & 40.3 &  32.0 & 34.3 & 30.8\\
\bottomrule
\end{tabular}}
\caption{Results on probing T5 and BART model with top-$k$ selection mechanism}
\label{tab:llm-results}
\end{table*}

\begin{table*}[t]
\resizebox{\textwidth}{!}{ 
\centering
\begin{tabular}{lccccc}
\toprule
Prompt Template & optimal $k$ & precision & recall &	max-F1 & p@1\\
\midrule
\x{} consists of \mask.& 	81.08& 	42.68& 	68.58& 	45.14& 	31.66\\ 
\x{} consists of \mask& 	251.54& 	22.73& 	26.18& 	20.08& 	11.05\\ 
\x{} consists of \mask{} element.& 	48.12& 	49.8& 	66.64& 	49.28& 	37.56\\ 
\x{} consists of \mask{} element& 	142.47& 	33.57& 	54.88& 	36.63& 	13.13\\ 
\x{} consists of \mask, which is an element.& 	33.44& 	67.84& 	72.2& 	64.11& 	68.00\\ 
\x{} consists of \mask, which is an element& 	39.69& 	60.18& 	70.56& 	57.03& 	54.27\\ 
The chemical compound \x{} consists of \mask.& 	23.11& 	50.65& 	73.09& 	52.23& 	42.00\\
\bottomrule
\end{tabular}}
\caption{Our proposed prompts for (chemical compound, has parts) relation.}
\label{tab:chemicalcompound}
\end{table*}

\begin{table*}[t]
\resizebox{\textwidth}{!}{ 
\centering
\begin{tabular}{lccccc}
\toprule
Prompt Template & optimal $k$ & precision & recall &	max-F1 & p@1\\
\midrule
\x{} shares border with \mask.& 	9.3& 	76.73& 	77.11& 	71.29& 	83.24\\ 
\x{} shares border with \mask& 	10.92& 	65.44& 	70.6& 	62.02& 	70.81\\ 
\x{} shares a border with \mask.& 	9.49& 	78.63& 	76.59& 	72.1& 	83.78\\ 
\x{} shares a border with \mask& 	11.62& 	61.24& 	69.24& 	58.22& 	65.95\\ 
\x{} borders \mask.& 	10.37& 	72.79& 	74.9& 	67.98& 	80.00\\ 
\x{} borders \mask& 	25.04& 	25.82& 	60.28& 	28.67& 	12.97\\ 
\x{} has borders with \mask.& 	9.54& 	79.26& 	75.14& 	71.41& 	82.70\\
\bottomrule
\end{tabular}}
\caption{Our proposed prompts for (country, shares borders) relation.}
\label{tab:country_borders}
\end{table*}

\begin{table*}[t]
\resizebox{\textwidth}{!}{ 
\centering
\begin{tabular}{lccccc}
\toprule
Prompt Template & optimal $k$ & precision & recall &	max-F1 & p@1\\
\midrule
The official language of \x{} is \mask.& 	18.18& 	92.58& 	80.43& 	83.54& 	91.84\\ 
The official language of \x{} is \mask& 	18.22& 	90.73& 	79.66& 	81.66& 	88.27\\ 
\mask{} is the official language of \x.& 	18.16& 	93.09& 	80.04& 	83.4& 	92.35\\ 
\mask{} is the official language of \x& 	18.2& 	92.15& 	80.17& 	83.1& 	91.33\\ 
\x{} has \mask{} as its official language.& 	18.3& 	87.26& 	79.62& 	79.51& 	83.16\\ 
\x{} has \mask{} as its official language& 	18.43& 	83.35& 	79.92& 	77.36& 	75.00\\ 
The official languages of \x{} are \mask.& 	18.19& 	92.08& 	80.3& 	83.02& 	90.31\\
\bottomrule
\end{tabular}}
\caption{Our proposed prompts for (country, has official language) relation.}
\label{tab:country_language}
\end{table*}

\begin{table*}[htbp]
\centering
\resizebox{0.95\textwidth}{!}{ 
\begin{tabular}{lccccc}
\toprule
Prompt Template & optimal $k$ & precision & recall &	max-F1 & p@1\\
\midrule
\x{} plays \mask.& 	83.77& 	52.59& 	60.22& 	49.03& 	49.00\\ 
\x{} plays \mask& 	56.31& 	31.27& 	66.57& 	32.52& 	20.50\\ 
\x{} plays \mask{} instrument.& 	17.48& 	36.88& 	76.01& 	42.07& 	24.00\\ 
\x{} plays \mask{} instrument& 	19.72& 	35.9& 	73.79& 	39.48& 	20.00\\ 
\x{} plays \mask{} musical instrument.& 	59.91& 	16.09& 	62.72& 	19.48& 	5.00\\ 
\x{} plays \mask{} musical instrument& 	71.2& 	11.67& 	61.89& 	15.84& 	2.50\\ 
\x{} plays \mask, which is an instrument.& 	14.12& 	65.16& 	74.57& 	61.24& 	61.50\\
\bottomrule
\end{tabular}}
\caption{Our proposed prompts for (person, plays an instrument) relation.}
\label{tab:person_instrument}
\end{table*}

\begin{table*}[t]
\centering
\resizebox{0.95\textwidth}{!}{ 
\begin{tabular}{lccccc}
\toprule
Prompt Template & optimal $k$ & precision & recall &	max-F1 & p@1\\
\midrule
\x{} speaks in \mask.& 	2.61& 	82.13& 	72.35& 	72.78& 	83.50\\ 
\x{} speaks in \mask& 	5.73& 	54.6& 	61.08& 	48.62& 	47.50\\ 
\x{} can speak in \mask.& 	2.56& 	81.63& 	69.37& 	70.76& 	82.50\\ 
\x{} can speak in \mask& 	7.26& 	63.35& 	66.07& 	57.14& 	60.50\\ 
\x{} communicates in \mask.& 	2.94& 	77.04& 	71.28& 	69.93& 	74.50\\ 
\x{} communicates in \mask& 	6.71& 	48.41& 	64.69& 	47.07& 	36.50\\ 
\x{} spoke in \mask.& 	2.59& 	82.62& 	70.24& 	71.32& 	86.00\\
\bottomrule
\end{tabular}}
\caption{Our proposed prompts for (person, speaks a language) relation.}
\label{tab:person_language}
\end{table*}

\begin{table*}[t]
\centering
\resizebox{0.85\textwidth}{!}{ 
\begin{tabular}{lccccc}
\toprule
Prompt Template & optimal $k$ & precision & recall &	max-F1 & p@1\\
\midrule
\x{} is a \mask{} by profession.& 	30.87& 	18.52& 	35.37& 	20.78& 	8.00\\ 
\x{} is a \mask{} by profession& 	28.39& 	20.08& 	34.17& 	21.23& 	13.00\\ 
\x{} is a \mask.& 	19.23& 	35.01& 	35.98& 	29.3& 	40.50\\ 
\x{} is a \mask& 	20.02& 	34.14& 	33.8& 	26.34& 	41.50\\ 
\x{} is a \mask, which is a profession.& 	23.15& 	30.17& 	36.72& 	26.81& 	31.00\\ 
\x{} is a \mask, which is a profession& 	17.87& 	35.11& 	31.63& 	26.43& 	40.00\\ 
\x's profession is \mask.& 	61.53& 	13.22& 	44.71& 	17.2& 	3.00\\
\bottomrule
\end{tabular}}
\caption{Our proposed prompts for (person, has an occupation) relation.}
\label{tab:person_profession}
\end{table*}

\begin{table*}[t]
\resizebox{\textwidth}{!}{ 
\centering
\begin{tabular}{lccccc}
\toprule
Prompt Template & optimal $k$ & precision & recall &	max-F1 & p@1\\
\midrule
\x{} shares border with \mask.& 	285.62& 	35.78& 	21.85& 	25.18& 	32.50\\ 
\x{} shares border with \mask& 	291.63& 	25.94& 	19.15& 	19.14& 	26.50\\ 
\x{} shares a border with \mask.& 	285.64& 	36.08& 	22.02& 	25.39& 	32.50\\ 
\x{} shares a border with \mask& 	296.5& 	22.06& 	18.2& 	17.32& 	20.00\\ 
\x{} borders \mask.& 	285.81& 	35.48& 	21.51& 	24.63& 	34.00\\ 
\x{} borders \mask& 	316.65& 	9.03& 	16.07& 	9.48& 	2.50\\ 
\x{} has borders with \mask.& 	286.25& 	32.6& 	21.67& 	24.1& 	30.00\\
\bottomrule
\end{tabular}}
\caption{Our proposed prompts for (state, shares border) relation.}
\label{tab:state_borders}
\end{table*}

\end{document}